\documentclass[sigconf]{acmart}
\AtBeginDocument{%
  \providecommand\BibTeX{{%
    \normalfont B\kern-0.5em{\scshape i\kern-0.25em b}\kern-0.8em\TeX}}}

\usepackage{xspace}
\usepackage{tikz}
\usepackage{graphicx}
\usepackage{amsmath}
\usepackage{xcolor}

\definecolor{hotpink}{HTML}{fe036a}
\newcommand{\tool}[1]{\textsc{Point \& Instruct\xspace{}}}

\usetikzlibrary{shapes.geometric} %

\newcommand{\thicksquarecyan}{
    \raisebox{-0.5mm}{
        \textcolor{cyan}{
            \begin{tikzpicture}
                \node[draw, thick, inner sep=0pt, minimum size=1em] {};
            \end{tikzpicture}
        }
    }
}
\newcommand{\starcyan}{
    \raisebox{-0.5mm}{
        \textcolor{cyan}{
            \begin{tikzpicture}
                \node[star, thick, star point ratio=2.25, minimum size=0.1cm, draw, scale=0.55] at (0,0) {};
            \end{tikzpicture}
        }
    }
}

\setcopyright{none}
\begin{document}

\title{\textsc{Point \& Instruct}: Enabling Precise Image Editing by Unifying Direct Manipulation and Text Instructions}

\author{Alec Helbling}
\affiliation{%
  \institution{Georgia Tech}
  \country{United States}}
\email{alechelbling@gatech.edu}
\author{Seongmin Lee}
\affiliation{%
  \institution{Georgia Tech}
  \country{United States}}
\email{seongmin@gatech.edu}
\author{Polo Chau}
\affiliation{%
  \institution{Georgia Tech}
  \country{United States}}
\email{polo@gatech.edu}

\begin{abstract}
    Machine learning has enabled the development of powerful systems capable of editing images from natural language instructions. However, in many common scenarios it is difficult for users to specify precise image transformations with text alone. For example, in an image with several dogs, it is difficult to select a particular dog and move it to a precise location.  Doing this with text alone would require a complex prompt that disambiguates the target dog and describes the destination. However, direct manipulation is well suited to visual tasks like selecting objects and specifying locations. We introduce Point and Instruct, a system for seamlessly combining familiar direct manipulation and textual instructions to enable precise image manipulation. With our system, a user can visually mark objects and locations, and reference them in textual instructions. This allows users to benefit from both the visual descriptiveness of natural language and the spatial precision of direct manipulation. 
\end{abstract}


\begin{teaserfigure}
    \centering
    \includegraphics[width=\textwidth]{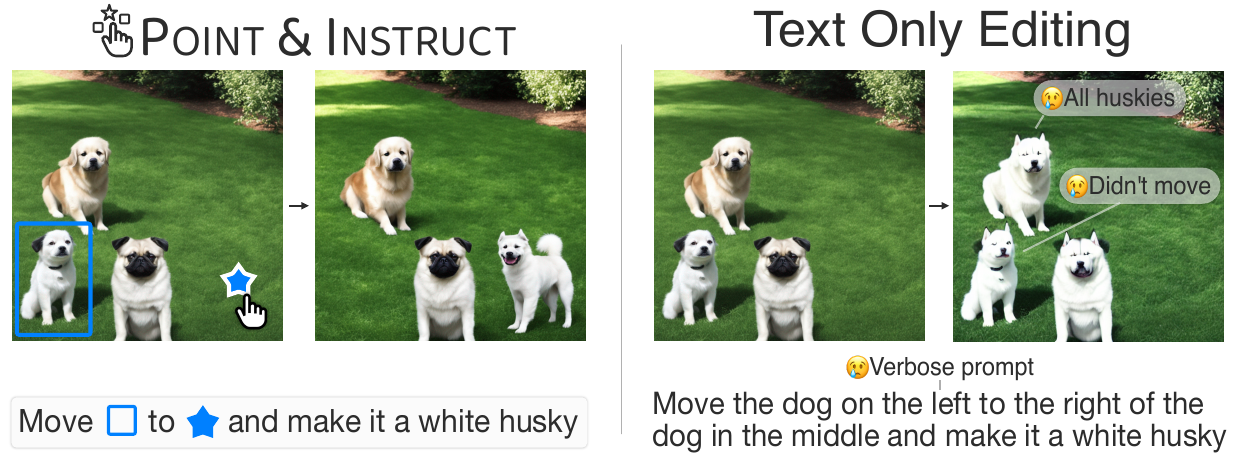}
    \vspace{-0.35in}
    \caption{\tool{} empowers users to specify image editing instructions that combine the expressively of natural language and the spatial precision of direct manipulation. We show an example of our method, which allows a user to move a particular dog to a precise location and change its appearance. (Left) A user can select which object in particular they wish to manipulate with a bounding box, and specify a location to move the object to with a star. These geometric shapes can be referenced in a natural language instruction symbolically and combined with language only instructions that specify changes to the appearance of objects. (Right) For comparison we show how the popular text-based editing system InstructPix2Pix \cite{brooks_instructpix2pix_2023} fails at this task. Not only does this system require a much more verbose query to convey the same image edit, but it also fails to move objects and fails to localize changes to the correct objects. }
    \label{fig:husky}
\end{teaserfigure}

\maketitle

\section{Introduction \label{section:intro}}

\begin{figure*}[t!]
  \begin{center}    
    \includegraphics[width=\textwidth]{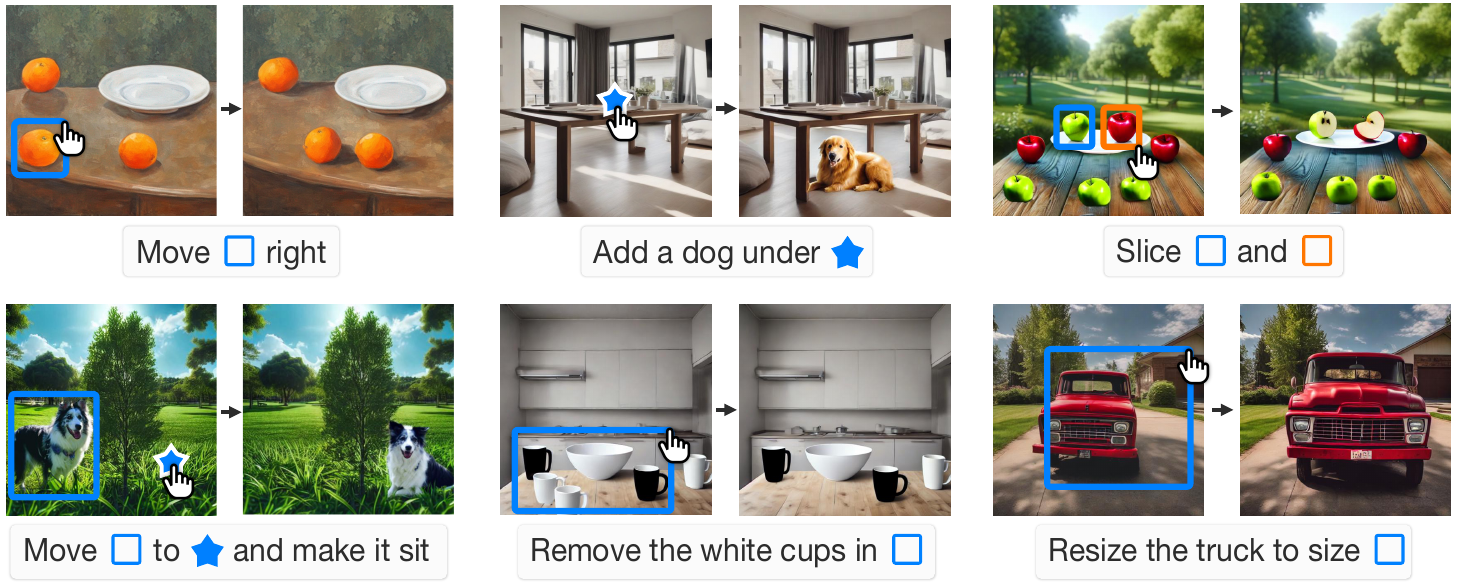}
  \end{center}
  \vspace{-0.1in}
  \caption{\textsc{Point \& Instruct} enables users to perform precise image manipulations that are difficult to do with text alone. A user can leverage familiar direct manipulation to specify regions or objects in an image, which can be referred to in text instructions. By combining direct manipulation and natural language based editing it becomes much easier for users to perform precise edits like: moving a particular object, adding an object in a specified location, or changing the appearance of an object.  
  }
  \label{fig:example-operations}
\end{figure*}

Recent advancements in machine learning have led to the development of powerful models capable of generating high-fidelity images from text descriptions \cite{rombach_high-resolution_2022, ramesh_zero-shot_2021, saharia_photorealistic_2022}. A similar class of models aim to empower users to edit existing images using text instructions \cite{brooks_instructpix2pix_2023, sheynin_emu_2023}. These systems allow users to draw on a large variety of descriptive vocabulary of noun phrases for identifying entities, adjectives for describing the appearance of objects, and verbs for describing visual transformations. However, many fine-grained image edits are difficult to specify with text alone. Figure \ref{fig:husky} shows an example scenario in which a user wishes to move a particular dog in an image with multiple dogs to a precise location and change its breed. To perform this task, text-only editing systems \cite{brooks_instructpix2pix_2023} require users  expend significant effort writing laboriously complex prompts, which even then might not elicit a correct edit from an image editing system. 

\begin{figure*}[t!]
    \centering
    \includegraphics[width=\textwidth]{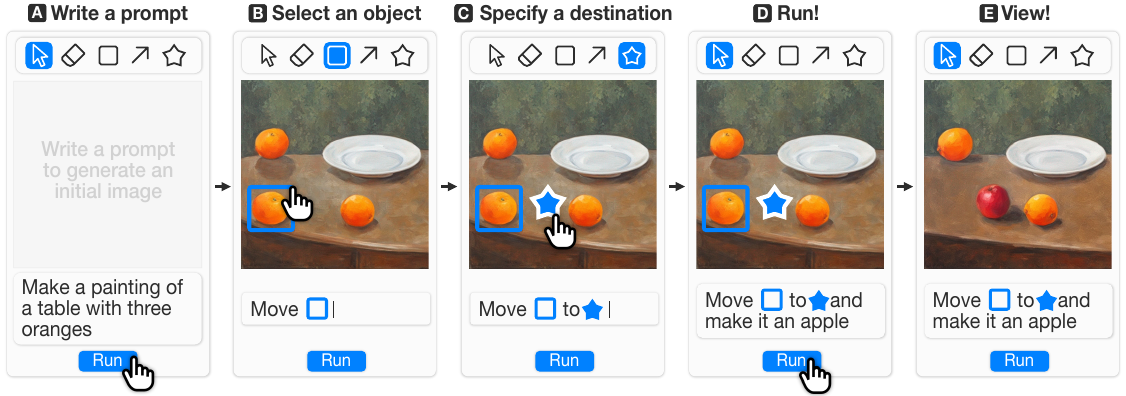}
    \vspace{-0.2in}
    \caption{\textsc{Point \& Instruct} harnesses the power of LLMs to process a variety of instructions and leverages visual information specified by simple geometric objects. The flexibility of text prompts can be seamlessly combined with familiar GUI elements, making it simple to understand and use. In our example use-case, a user would (A) upload an existing image or write a text prompt to generate an image, (B) select object(s) with a bounding box specified through direct manipulation, (C) specify another location to move an object to with a bounding box or star, (D) click enter or a button to run the generation process, and finally (E) view the generated image. }
    \label{fig:interface}
\end{figure*}

Direct manipulation has been a staple design principle in interactive applications, 
enabling users to more easily learn how an interface works because it builds upon a person's preexisting physical intuition \cite{shneiderman1997direct, shneiderman1983direct, hutchins_direct_1985, shneiderman_future_1982}.
It is particularly well suited to image editing tasks like isolating particular regions in an image to edit because a user can very easily click on an object or region. 
However, direct manipulation alone does not allow users to leverage the expressiveness of natural language to specify complex visual transformations (e.g., changing a dog breed). This motivates us to combine direct manipulation and natural language instructions, forming a multi-modal interface for image editing. In this ongoing work, our primary contributions are:
\begin{enumerate}
    \item \textbf{\tool{}} (Figure \ref{fig:husky}), a system that enables users to make precise image edits by unifying familiar direct manipulation \cite{hutchins_direct_1985, shneiderman_future_1982} and natural language instructions \cite{perrault_chapter_1988}. 
    \tool{} takes the form of a web-based interactive tool for specifying multimodal editing interactions (see Figure \ref{fig:interface}). 
    With \textsc{Point \& Instruct}, a user can leverage the spatial precision of visual inputs (e.g., points and bounding boxes), and the visual descriptiveness of textual instructions (e.g., ``make the dog brown'') to specify fine-grained manipulations to images. Figure \ref{fig:example-operations} demonstrates  precise manipulations enabled by \tool{}.
    \item \textbf{A Novel, General Framework for Combining Direct Manipulation and Textual Instructions for Visual Manipulation}. We frame the task of following visual instructions for image editing as a natural language generation task, enabling the application of  LLMs to a visual manipulation task. We serialize both the layout of an image and the visual elements (e.g., bounding boxes) of user instructions into a textual format, allowing an LLM to process them. We believe that this framework can extend beyond image editing to other visual tasks, like web design or presentation construction. 

\end{enumerate}

\section{Background and Related Works}

\textbf{Text-based Image Editing.} Recent advancements in machine learning have led to the development of systems capable of high-fidelity image synthesis \cite{rombach_high-resolution_2022, ramesh_zero-shot_2021, goodfellow_generative_2014}. Generating images from text descriptions \cite{rombach_high-resolution_2022} has become a popular technique due to the visual descriptiveness of natural language, and the scalability of collecting image-caption pairs from the internet \cite{schuhmann2022laion5b}. In addition to generating images from text descriptions, there has been tremendous progress in text-based image editing \cite{brooks_instructpix2pix_2023, hertz_prompt--prompt_2022, parmar_zero-shot_2023}. However, both text-to-image generation and text based editing systems struggle with tasks like generating consistent recurring subjects \cite{ruiz_dreambooth_2023}, satisfying spatial relationships specified by text prompts \cite{liu_compositional_2022, hu_tifa_2023}, and are prone to generating unwanted inappropriate content \cite{schramowski_safe_2023}. Furthermore, as we discuss in Section \ref{section:intro}, natural language is ill-suited for precise image editing tasks that require a user to disambiguate a particular object or specify a precise location. Our work attempts to solve this problem by combining text-based instruction following with direct manipulation.

\textbf{Layout-Based Image Generation.} A key problem with text-only image generation systems is that it is quite difficult or even impossible to convey precise spatial positions of various objects in an image with text alone. This has motivated the development of grounded image generation techniques \cite{chen_training-free_2023, li_gligen_2023} which allow users to control the positions of objects in a generated image. Another line of work aims to generate such layouts from text descriptions \cite{lian_llm-grounded_2023, feng_layoutgpt_2023}. These works make the observation that it is possible to serialize an image layout into a JSON-like textual format, which can be readily processed by LLMs. This follows a broader theme in the field of applying language-only LLMs to vision tasks through visual instruction tuning \cite{liu_visual_2023}. Our work builds upon the observations of this line of work, demonstrating that it is possible to use LLMs to process multi-modal instructions for the application of image editing.

\textbf{Direct Manipulation and Natural Language Interfaces.} Direct manipulation interfaces have long been a core component of many graphical user interfaces \cite{shneiderman_1980, hutchins_direct_1985, shneiderman_future_1982}. Likewise, building flexible natural language interfaces has long been of interest in the HCI community \cite{perrault_chapter_1988}. Several existing works attempt to demonstrate how combining direct manipulation and natural language interfaces helps overcome the limitations of both mediums \cite{cohen_natural_language_1992, cohen_synergistic_1989}. Motivated by this, we build a system for multi-modal interaction that combines direct manipulation with natural language interfaces for the application of image editing.

\section{System Design and Implementation}

\begin{figure*}[t]
    \centering
    \includegraphics[width=0.9\textwidth]{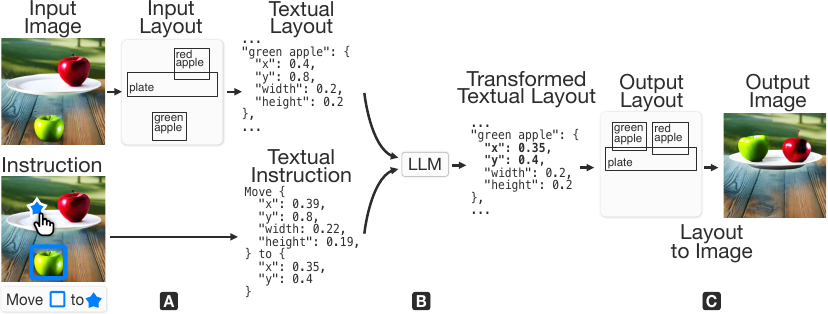}
    \caption{\textsc{Point \& Instruct} casts the problem of image editing as a natural language generation task. (A) The input image and instruction are serialized into a textual form, and (B) an LLM accepts the input layout and instruction and produces a transformed layout. Finally, (C) a layout-to-image generation system is used to generate an edited image from the transformed layout. }
    \label{fig:system}
\end{figure*}

\textsc{Point \& Instruct} is a system for image editing that seamlessly combines direct manipulation and natural language instructions. Our system is composed of several key components: a \textit{multimodal user interface} (Figure \ref{fig:interface}) that combines direct manipulation and natural language, a \textit{multimodal instruction following module} (see Figure \ref{fig:system}a) that is responsible for processing a users multimodal instructions, and a \textit{layout-based image generation} component that leverages diffusion models to generate an image from a specified layout (see Figure \ref{fig:system}b). 

\subsection{User Interface \label{sec:interface}}

\textbf{Interface Design.} We aimed to develop a simple user interface that requires little familiarity with sophisticated graphic design software (see Figure \ref{fig:interface}). Rather than having a complex WIMP-style interface \cite{van_dam_post-wimp_1997} with many layers of dropdown menus, we simply have five tools organized in a toolbar, an interactive canvas, and a instruction input form. The tools supported by the interface are: a \textit{select} operation for selecting drawn objects, an \textit{erase} operation for removing the objects, a \textit{bounding box} tool, an \textit{arrow} tool for drawing arrows, and a \textit{star} tool for specifying 2D points on the canvas. Despite its simplicity, users can perform a wide array of image manipulations (see Figure \ref{fig:example-operations}). Rather than having a pre-specified tool for primitive manipulation operations like \textit{moving} an object, \textit{adding} an object, or \textit{changing the appearance of an object}, we allow user so specify these transformations using flexible language instructions. Users can leverage direct manipulation to easily specify geometric objects (e.g., bounding boxes) which are used as ``arguments'' for these natural language editing instructions. As the user draws each of these objects, they appear in the instruction text box with reference symbols as if they were words in the instruction text box. 

\textbf{Usage Scenario.} We present an example usage scenario in Figure \ref{fig:interface}. Let Alice be a typical user who wishes to generate a simple painting. She could write a prompt like ``make a painting of a table with three oranges'' and click run, and \textsc{Point \& Instruct} will then generate an image that matches that description. If Alice does not like every detail of that image she can perform tasks like moving objects or changing their appearance. Say Alice wants to move one of the oranges in the image. She can select a bounding box around an orange she wishes to move by selecting the bounding box tool and using her cursor. Further, she can specify the place she wants to move the object to using a star or a bounding box. A limitation of direct manipulation alone is that it would be difficult for Alice to make a very specific change in the appearance of an object, like turning it into an apple. It is impractical for a direct manipulation interface to have a pre-specified set of tools for every possible transformation. However, with our tool, Alice can simply say ``make\thicksquarecyan\hspace{-0.05in} an apple'' to turn a selected object into an apple. 

\subsection{Multi-modal Instruction Following} \label{multi-modal-instructions}

Our goal is to take an input image and a multi-modal instruction and produce a transformed image that resolves the instruction. Instead of directly transforming images using a diffusion model we instead manipulate an intermediate representation of an image in the form of a spatial layout of objects specified by bounding boxes and text descriptions. 
We observe that it is possible to serialize multi-modal instructions composed of both text and simple geometric objects like bounding boxes into a textual form, allowing them to be processed by an LLM. Casting the problem in this way allows us to drastically simplify the problem of image manipulation, operating in image layout space rather than pixel space. Furthermore, we inherit several key advantages of LLMs like their ability to learn from just a few in-context examples and generalize to unseen vocabulary. 

Using our interface (see Section \ref{sec:interface}) a user can specify multi-modal instructions composed of both natural language and geometric objects like bounding boxes. 
Simple geometric objects like points can be represented as ``\texttt{\{x: 144, y:132\}}'' and a bounding box can be represented as ``\texttt{\{x: 150, y: 400, width: 100, height: 100\}}''. This allows us to convert multi-modal instructions into a textual form that can be readily processed by LLMs. 
Suppose a user specifies a bounding box\thicksquarecyan\hspace{-0.05in}and an instruction ``move\thicksquarecyan\hspace{-0.05in} to \starcyan\hspace{-0.05in}''. 
We represent this instruction as text like ``move \texttt{\{x: 150, y:400, width: 100, height: 100\}} to \texttt{\{x: 144, y: 132\}}'' (see Figure \ref{fig:system}A). 
By feeding an instruction into an LLM we can generate a transformed layout (Figure \ref{fig:system}B), which can then be given to a layout-to-image generation system (Figure \ref{fig:system}C) like \textsc{GLIGEN} \cite{li_gligen_2023}. 
Furthermore, this framing supports more than just spatial transformations. Say a user writes an instruction ``make the dog in \thicksquarecyan black'', an LLM can change the corresponding dog's prompt from ``a white dog'' to ``a black dog" in the textual layout representation. This overall framework is shown in Figure \ref{fig:system}. 

\begin{figure*}
    \centering
    \includegraphics[width=\textwidth]{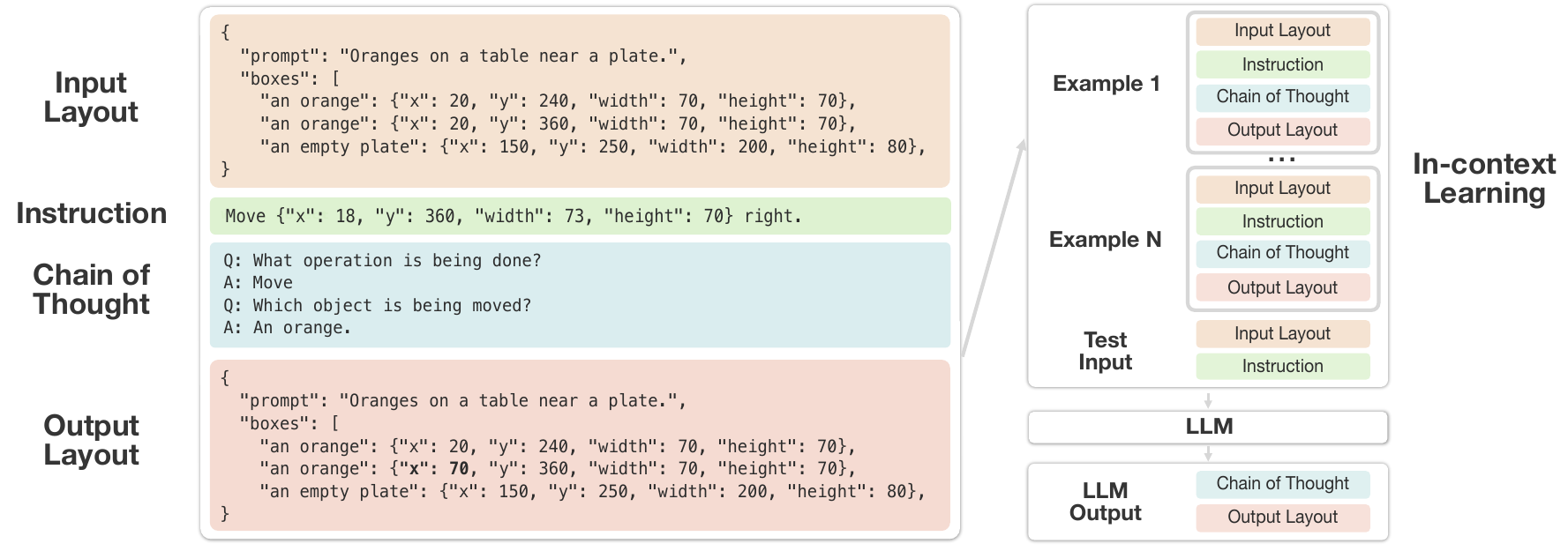}
    \caption{We leverage in-context learning to take advantage of the few-shot generalization capabilities of LLMs. We place a relatively small number ($\approx 15$) examples for our task in the context of an LLM. Each example contains (a) a serialized image layout, (b) a serialized instruction, (c) a chain of thought composed of multiple task-relevant questions meant to assist the LLM by providing it with additional context, and (d) an annotated layout specifying the relevant transformation. At inference time we place an input image layout and instruction after the in-context examples. }
    \label{fig:in-context-learning}
\end{figure*}

\textbf{In-context Learning and Chain of Thought Prompting.}
Instead of performing data intensive and computationally expensive fine-tuning of an underlying language model we leverage in-context learning. In-context learning \cite{min_metaicl_2022} is a powerful technique which allows us to put examples of a task into the context window of a large-language model at inference time. This has been shown to allow for few-shot generalization to a new task. Figure \ref{fig:in-context-learning} shows an example of the structure of data for our task. We provide a textual specification of an image layout, followed by a textual multi-modal instruction. We then leverage chain-of-thought prompting by presenting the language model with intermediate questions which has been shown to improve performance on a wide variety of tasks \cite{wei_chain--thought_2023}, followed by an edited output layout. The authors hand annotated a relatively small number ($\approx 20$) of examples composed of input layouts, instructions, output layouts, and chain of thought prompts. We place multiple instances of these examples into the context window of the LLM, and at test time have an input layout and instruction. The LLM uses all of these examples to ``autocomplete'' the chain of thought and manipulation of the unseen layout and instruction. 

\subsection{Layout Based Image Generation} 

The final key component of our system is a layout based manipulation system. Existing methods demonstrate that it is possible to generate images that adhere to user specified layouts \cite{li_gligen_2023, chen_training-free_2023}. However, in our setting we have an additional constraint, which is that we want to modify an existing image to have an altered layout and content. We require a system which can accept an input image and perform tasks like moving an object while retaining its core appearance. For our experiments, we leverage a combination of training-free layout guidance \cite{chen_training-free_2023} and GLIGEN \cite{li_gligen_2023} to conditionally generate images given a text layout. We inpaint each object onto a fixed background image which we also generate using a diffusion model. By inpainting onto a fixed background image we can move around objects without making unwanted changes to the scenery. However, a limitation of our system is that the inpainted objects do not always have a perfectly consistent appearance across different generations. One solution to this is to fine-tune a diffusion model with images of a particular subject \cite{ruiz_dreambooth_2023, li_dreamedit_2023}, binding that subject to a particular token which can be used to generate new images of the subject. We demonstrate this approach in the bottom left example of Figure \ref{fig:example-operations}. However, this approach is limited in its scalability as it requires fine-tuning a model for each subject. Consistent subject-driven generation across multiple images is a open problem, and is an avenue for future work.

\subsection{Implementation Details}

To implement the user interface of \tool{} we use ReactJS \cite{react2024} with a package called Tldraw \cite{tldraw2024} to implement the interactive canvas. The front end connects to a Python Flask backend server that interfaces with the multi-modal instruction following system described in Section \ref{multi-modal-instructions}. For our LLM component we leverage the GPT 3.5-Turbo API \cite{brown_language_2020} with in-context learning, although in principle our framework could be used with a number of LLMs. For layout based image generation we leverage a both GLIGEN\cite{li_gligen_2023} and LMD+\cite{lian_llm-grounded_2023}. These models are built on top of Stable Diffusion \cite{rombach_high-resolution_2022}. For certain generations we leverage subject based fine-tuning \cite{ruiz_dreambooth_2023} to improve subject consistency, and leverage DiffEdit \cite{couairon_diffedit_2022} to modify existing generations with altered subjects. This is similar to an approach done in \cite{li_dreamedit_2023}. 

\section{Results and Comparison to Existing Work}

\begin{figure*}
    \centering
    \includegraphics[width=\textwidth]{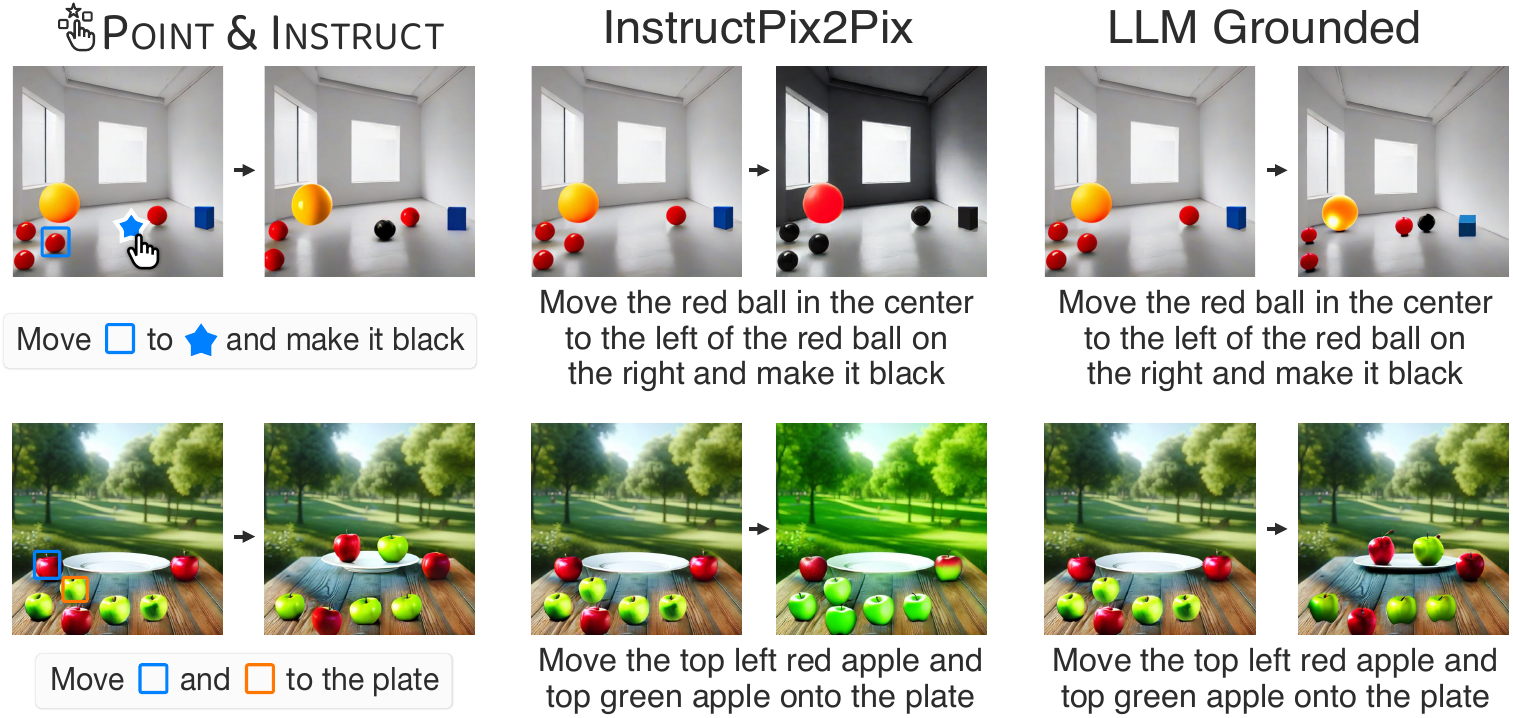}
    \caption{Our approach enables a user to disambiguate a particular object from other similar objects, move it, and change its appearance. In contrast, text-only editing approaches like \textsc{InstructPix2Pix} \cite{brooks_instructpix2pix_2023} and \textsc{LLM Grounded Diffusion} \cite{lian_llm-grounded_2023} fail to localize the manipulation to the correct object, despite requiring a much longer and more difficult to write edit instruction. }
    \label{fig:comparison}
\end{figure*}

We beleive it is worth highlighting scenarios where \tool{} especially shines when compared to existing text only image editing methods like \textsc{InstructPix2Pix} \cite{brooks_instructpix2pix_2023} and \textsc{LLM Grounded Diffusion} (LGD) \cite{lian_llm-grounded_2023}. For the sake of comparison, for both our approach and \cite{lian_llm-grounded_2023} we leveraged the same layout-based generation pipeline, namely a GLIGEN inpainting pipeline \cite{li_gligen_2023}. Figure \ref{fig:comparison} compares \tool{} with these two baselines on two separate editing tasks. The first task shows a user moving a particular red ball in a complex image with several red balls. With \tool{} a user can simply select the ball with a bounding box, specify its target location, and use text to describe an appearance transformation like ``make it black''. Text-only models however require much more complex queries to precisely disambiguate the correct red ball and specify the destination location. \textsc{InstructPix2Pix} fails to move the target object and fails to isolate the ``make it black'' instruction to the correct object. LGD successfully moves the correct ball, but it moves it to an incorrect location.  In the second example, a user wishes to move two selected apples onto a plate. Our approach allows a user to perform this operation successfully with a very concise instruction. However, \textsc{InstructPix2Pix} again fails to move objects, and unnecessarily changes the color of some of the apples to green. LGD is able to successfully resolve the instruction, but the instruction is much more complex and tedious to write than our system's multi-modal instruction.

\section{Ongoing Work}

\subsection{Human Evaluation}

In order to understand the usability of our user interface and the usefulness of the variety of operations offered by \tool{} we will conduct two user studies. We will deploy our system in a Docker container environment and host it on Amazon Web Services. 
Our first user study will focus on the usability of our interactive web interface (Figure \ref{fig:interface}). 
The simplicity of our tool makes it easy to learn for users without any particular familiarity with design software. 
We plan to conduct a large-scale user study with 100 participants from Amazon Mechanical Turk. 
Our objective is to evaluate the usability of our tool, such as how easy it is for participants to take an input image and manipulate it to resemble a given target image. 
For each trial, participants will be provided with an ``original'' input image generated using a text prompt collected from the DiffusionDB dataset \cite{wang2023diffusiondb}. We will manually manipulate each of these images in certain fine-grained ways (e.g., moving objects, changing the appearance of particular objects, etc. ), producing a ground truth target image for each corresponding input image. The participants' task will be to manipulate the input image to look as close as possible to the target image. 
Before engaging in the editing task, each participant will watch a brief one-minute  video tutorial about how to use the interface. Information like which instructions a user wrote, how many instructions they wrote, their mouse interactions, and the intermediate generated images will be collected. Participants will be presented with a brief survey of questions on a Likert scale focusing on the usability of the interface, its learnability, their enjoyment, the speed and responsiveness, and their satisfaction with their edited image. Separate participants will be recruited from Mechanical Turk to assess how well the final images produced by each user match the ground truth target image presented to each user. 

A second user study will be conducted to assess the benefits of \tool{} over existing text-only editing systems \cite{brooks_instructpix2pix_2023, lian_llm-grounded_2023}. The structure of this study will be similar to the first one, but we will also have users use a text-only interface for manipulating images. We will likewise survey each user on factors like the usability of text only interfaces, their satisfaction with the final image, and present the final image to independent annotators who assess the similarity of the user generated image to the ground truth target image.

\subsection{Evaluating and Improving the Performance of Multi-modal Editing Instructions}

A challenge that we have faced in the development of \textsc{Point \& Instruct} is evaluating our LLM based instruction following component. It is important to ensure that multi-modal instructions correctly manipulate the layout of an image. This task is difficult to automate because there are certain classes of transformations, like ``move the ball to the right'', that have an infinite possible number of valid solutions. In this example, there may be many locations in an image that may be to the right of the original location. We will perform human evaluation to analyze the efficacy of our system on these types of transformations. A promising avenue for future work is systems that can evaluate the efficacy of multi-modal instruction following in an automated fashion.

\section{Conclusion}

We introduce \textsc{Point \& Instruct}, a novel system enabling user's to make precise manipulations to images by combining natural language instructions and direct manipulation. We are working to further evaluate the efficacy of \textsc{Point \& Instruct} through extensive user studies, and systematically evaluate the accuracy of our system for multi-modal image editing.

\section{Acknowledgements}

This work is supported by DARPA GARD. 

\bibliographystyle{ACM-Reference-Format}
\bibliography{references}


\begin{thebibliography}{35}


\ifx \showCODEN    \undefined \def \showCODEN     #1{\unskip}     \fi
\ifx \showDOI      \undefined \def \showDOI       #1{#1}\fi
\ifx \showISBNx    \undefined \def \showISBNx     #1{\unskip}     \fi
\ifx \showISBNxiii \undefined \def \showISBNxiii  #1{\unskip}     \fi
\ifx \showISSN     \undefined \def \showISSN      #1{\unskip}     \fi
\ifx \showLCCN     \undefined \def \showLCCN      #1{\unskip}     \fi
\ifx \shownote     \undefined \def \shownote      #1{#1}          \fi
\ifx \showarticletitle \undefined \def \showarticletitle #1{#1}   \fi
\ifx \showURL      \undefined \def \showURL       {\relax}        \fi
\providecommand\bibfield[2]{#2}
\providecommand\bibinfo[2]{#2}
\providecommand\natexlab[1]{#1}
\providecommand\showeprint[2][]{arXiv:#2}

\bibitem[rea(2024)]%
        {react2024}
 \bibinfo{year}{2024}\natexlab{}.
\newblock \bibinfo{title}{React}.
\newblock \bibinfo{howpublished}{\url{https://react.dev/}}.
\newblock
\newblock
\shownote{Accessed: 2024-01-25}.


\bibitem[tld(2024)]%
        {tldraw2024}
 \bibinfo{year}{2024}\natexlab{}.
\newblock \bibinfo{title}{tldraw}.
\newblock \bibinfo{howpublished}{\url{https://tldraw.dev/}}.
\newblock
\newblock
\shownote{Accessed: 2024-01-25}.


\bibitem[Brooks et~al\mbox{.}(2023)]%
        {brooks_instructpix2pix_2023}
\bibfield{author}{\bibinfo{person}{Tim Brooks}, \bibinfo{person}{Aleksander
  Holynski}, {and} \bibinfo{person}{Alexei~A. Efros}.}
  \bibinfo{year}{2023}\natexlab{}.
\newblock \bibinfo{title}{{InstructPix2Pix}: {Learning} to {Follow} {Image}
  {Editing} {Instructions}}.
\newblock
\newblock
\urldef\tempurl%
\url{https://doi.org/10.48550/arXiv.2211.09800}
\showDOI{\tempurl}
\newblock
\shownote{arXiv:2211.09800 [cs]}.


\bibitem[Brown et~al\mbox{.}(2020)]%
        {brown_language_2020}
\bibfield{author}{\bibinfo{person}{Tom~B. Brown}, \bibinfo{person}{Benjamin
  Mann}, \bibinfo{person}{Nick Ryder}, \bibinfo{person}{Melanie Subbiah},
  \bibinfo{person}{Jared Kaplan}, \bibinfo{person}{Prafulla Dhariwal},
  \bibinfo{person}{Arvind Neelakantan}, \bibinfo{person}{Pranav Shyam},
  \bibinfo{person}{Girish Sastry}, \bibinfo{person}{Amanda Askell},
  \bibinfo{person}{Sandhini Agarwal}, \bibinfo{person}{Ariel Herbert-Voss},
  \bibinfo{person}{Gretchen Krueger}, \bibinfo{person}{Tom Henighan},
  \bibinfo{person}{Rewon Child}, \bibinfo{person}{Aditya Ramesh},
  \bibinfo{person}{Daniel~M. Ziegler}, \bibinfo{person}{Jeffrey Wu},
  \bibinfo{person}{Clemens Winter}, \bibinfo{person}{Christopher Hesse},
  \bibinfo{person}{Mark Chen}, \bibinfo{person}{Eric Sigler},
  \bibinfo{person}{Mateusz Litwin}, \bibinfo{person}{Scott Gray},
  \bibinfo{person}{Benjamin Chess}, \bibinfo{person}{Jack Clark},
  \bibinfo{person}{Christopher Berner}, \bibinfo{person}{Sam McCandlish},
  \bibinfo{person}{Alec Radford}, \bibinfo{person}{Ilya Sutskever}, {and}
  \bibinfo{person}{Dario Amodei}.} \bibinfo{year}{2020}\natexlab{}.
\newblock \bibinfo{title}{Language {Models} are {Few}-{Shot} {Learners}}.
\newblock
\newblock
\urldef\tempurl%
\url{https://doi.org/10.48550/arXiv.2005.14165}
\showDOI{\tempurl}
\newblock
\shownote{arXiv:2005.14165 [cs]}.


\bibitem[Chen et~al\mbox{.}(2023)]%
        {chen_training-free_2023}
\bibfield{author}{\bibinfo{person}{Minghao Chen}, \bibinfo{person}{Iro Laina},
  {and} \bibinfo{person}{Andrea Vedaldi}.} \bibinfo{year}{2023}\natexlab{}.
\newblock \bibinfo{title}{Training-{Free} {Layout} {Control} with
  {Cross}-{Attention} {Guidance}}.
\newblock
\newblock
\urldef\tempurl%
\url{https://doi.org/10.48550/arXiv.2304.03373}
\showDOI{\tempurl}
\newblock
\shownote{arXiv:2304.03373 [cs]}.


\bibitem[Cohen(1992)]%
        {cohen_natural_language_1992}
\bibfield{author}{\bibinfo{person}{Philip~R. Cohen}.}
  \bibinfo{year}{1992}\natexlab{}.
\newblock \showarticletitle{The role of natural language in a multimodal
  interface}. In \bibinfo{booktitle}{\emph{Proceedings of the 5th Annual ACM
  Symposium on User Interface Software and Technology}} (Monteray, California,
  USA) \emph{(\bibinfo{series}{UIST '92})}. \bibinfo{publisher}{Association for
  Computing Machinery}, \bibinfo{address}{New York, NY, USA},
  \bibinfo{pages}{143–149}.
\newblock
\showISBNx{0897915496}
\urldef\tempurl%
\url{https://doi.org/10.1145/142621.142641}
\showDOI{\tempurl}


\bibitem[Cohen et~al\mbox{.}(1989)]%
        {cohen_synergistic_1989}
\bibfield{author}{\bibinfo{person}{P.~R. Cohen}, \bibinfo{person}{M.
  Dalrymple}, \bibinfo{person}{D.~B. Moran}, \bibinfo{person}{F.~C. Pereira},
  {and} \bibinfo{person}{J.~W. Sullivan}.} \bibinfo{year}{1989}\natexlab{}.
\newblock \showarticletitle{Synergistic use of direct manipulation and natural
  language}. In \bibinfo{booktitle}{\emph{Proceedings of the SIGCHI Conference
  on Human Factors in Computing Systems}} \emph{(\bibinfo{series}{CHI '89})}.
  \bibinfo{publisher}{Association for Computing Machinery},
  \bibinfo{address}{New York, NY, USA}, \bibinfo{pages}{227–233}.
\newblock
\showISBNx{0897913019}
\urldef\tempurl%
\url{https://doi.org/10.1145/67449.67494}
\showDOI{\tempurl}


\bibitem[Couairon et~al\mbox{.}(2022)]%
        {couairon_diffedit_2022}
\bibfield{author}{\bibinfo{person}{Guillaume Couairon}, \bibinfo{person}{Jakob
  Verbeek}, \bibinfo{person}{Holger Schwenk}, {and} \bibinfo{person}{Matthieu
  Cord}.} \bibinfo{year}{2022}\natexlab{}.
\newblock \bibinfo{title}{{DiffEdit}: {Diffusion}-based semantic image editing
  with mask guidance}.
\newblock
\newblock
\urldef\tempurl%
\url{https://doi.org/10.48550/arXiv.2210.11427}
\showDOI{\tempurl}
\newblock
\shownote{arXiv:2210.11427 [cs]}.


\bibitem[Feng et~al\mbox{.}(2023)]%
        {feng_layoutgpt_2023}
\bibfield{author}{\bibinfo{person}{Weixi Feng}, \bibinfo{person}{Wanrong Zhu},
  \bibinfo{person}{Tsu-jui Fu}, \bibinfo{person}{Varun Jampani},
  \bibinfo{person}{Arjun Akula}, \bibinfo{person}{Xuehai He},
  \bibinfo{person}{Sugato Basu}, \bibinfo{person}{Xin~Eric Wang}, {and}
  \bibinfo{person}{William~Yang Wang}.} \bibinfo{year}{2023}\natexlab{}.
\newblock \bibinfo{title}{{LayoutGPT}: {Compositional} {Visual} {Planning} and
  {Generation} with {Large} {Language} {Models}}.
\newblock
\newblock
\urldef\tempurl%
\url{https://doi.org/10.48550/arXiv.2305.15393}
\showDOI{\tempurl}
\newblock
\shownote{arXiv:2305.15393 [cs]}.


\bibitem[Goodfellow et~al\mbox{.}(2014)]%
        {goodfellow_generative_2014}
\bibfield{author}{\bibinfo{person}{Ian~J. Goodfellow}, \bibinfo{person}{Jean
  Pouget-Abadie}, \bibinfo{person}{Mehdi Mirza}, \bibinfo{person}{Bing Xu},
  \bibinfo{person}{David Warde-Farley}, \bibinfo{person}{Sherjil Ozair},
  \bibinfo{person}{Aaron Courville}, {and} \bibinfo{person}{Yoshua Bengio}.}
  \bibinfo{year}{2014}\natexlab{}.
\newblock \bibinfo{title}{Generative {Adversarial} {Networks}}.
\newblock
\newblock
\urldef\tempurl%
\url{https://doi.org/10.48550/arXiv.1406.2661}
\showDOI{\tempurl}
\newblock
\shownote{arXiv:1406.2661 [cs, stat]}.


\bibitem[Hertz et~al\mbox{.}(2022)]%
        {hertz_prompt--prompt_2022}
\bibfield{author}{\bibinfo{person}{Amir Hertz}, \bibinfo{person}{Ron Mokady},
  \bibinfo{person}{Jay Tenenbaum}, \bibinfo{person}{Kfir Aberman},
  \bibinfo{person}{Yael Pritch}, {and} \bibinfo{person}{Daniel Cohen-Or}.}
  \bibinfo{year}{2022}\natexlab{}.
\newblock \bibinfo{title}{Prompt-to-{Prompt} {Image} {Editing} with {Cross}
  {Attention} {Control}}.
\newblock
\newblock
\urldef\tempurl%
\url{https://doi.org/10.48550/arXiv.2208.01626}
\showDOI{\tempurl}
\newblock
\shownote{arXiv:2208.01626 [cs]}.


\bibitem[Hu et~al\mbox{.}(2023)]%
        {hu_tifa_2023}
\bibfield{author}{\bibinfo{person}{Yushi Hu}, \bibinfo{person}{Benlin Liu},
  \bibinfo{person}{Jungo Kasai}, \bibinfo{person}{Yizhong Wang},
  \bibinfo{person}{Mari Ostendorf}, \bibinfo{person}{Ranjay Krishna}, {and}
  \bibinfo{person}{Noah~A. Smith}.} \bibinfo{year}{2023}\natexlab{}.
\newblock \bibinfo{title}{{TIFA}: {Accurate} and {Interpretable}
  {Text}-to-{Image} {Faithfulness} {Evaluation} with {Question} {Answering}}.
\newblock
\newblock
\urldef\tempurl%
\url{https://doi.org/10.48550/arXiv.2303.11897}
\showDOI{\tempurl}
\newblock
\shownote{arXiv:2303.11897 [cs]}.


\bibitem[Hutchins et~al\mbox{.}(1985)]%
        {hutchins_direct_1985}
\bibfield{author}{\bibinfo{person}{Edwin~L. Hutchins},
  \bibinfo{person}{James~D. Hollan}, {and} \bibinfo{person}{Donald~A. Norman}.}
  \bibinfo{year}{1985}\natexlab{}.
\newblock \showarticletitle{Direct {Manipulation} {Interfaces}}.
\newblock \bibinfo{journal}{\emph{Human–Computer Interaction}}
  \bibinfo{volume}{1}, \bibinfo{number}{4} (\bibinfo{date}{Dec.}
  \bibinfo{year}{1985}), \bibinfo{pages}{311--338}.
\newblock
\showISSN{0737-0024}
\urldef\tempurl%
\url{https://doi.org/10.1207/s15327051hci0104_2}
\showDOI{\tempurl}
\newblock
\shownote{Publisher: Taylor \& Francis \_eprint:
  https://doi.org/10.1207/s15327051hci0104\_2}.


\bibitem[Li et~al\mbox{.}(2023a)]%
        {li_dreamedit_2023}
\bibfield{author}{\bibinfo{person}{Tianle Li}, \bibinfo{person}{Max Ku},
  \bibinfo{person}{Cong Wei}, {and} \bibinfo{person}{Wenhu Chen}.}
  \bibinfo{year}{2023}\natexlab{a}.
\newblock \bibinfo{title}{{DreamEdit}: {Subject}-driven {Image} {Editing}}.
\newblock
\newblock
\urldef\tempurl%
\url{https://doi.org/10.48550/arXiv.2306.12624}
\showDOI{\tempurl}
\newblock
\shownote{arXiv:2306.12624 [cs]}.


\bibitem[Li et~al\mbox{.}(2023b)]%
        {li_gligen_2023}
\bibfield{author}{\bibinfo{person}{Yuheng Li}, \bibinfo{person}{Haotian Liu},
  \bibinfo{person}{Qingyang Wu}, \bibinfo{person}{Fangzhou Mu},
  \bibinfo{person}{Jianwei Yang}, \bibinfo{person}{Jianfeng Gao},
  \bibinfo{person}{Chunyuan Li}, {and} \bibinfo{person}{Yong~Jae Lee}.}
  \bibinfo{year}{2023}\natexlab{b}.
\newblock \bibinfo{title}{{GLIGEN}: {Open}-{Set} {Grounded} {Text}-to-{Image}
  {Generation}}.
\newblock
\newblock
\urldef\tempurl%
\url{https://doi.org/10.48550/arXiv.2301.07093}
\showDOI{\tempurl}
\newblock
\shownote{arXiv:2301.07093 [cs]}.


\bibitem[Lian et~al\mbox{.}(2023)]%
        {lian_llm-grounded_2023}
\bibfield{author}{\bibinfo{person}{Long Lian}, \bibinfo{person}{Boyi Li},
  \bibinfo{person}{Adam Yala}, {and} \bibinfo{person}{Trevor Darrell}.}
  \bibinfo{year}{2023}\natexlab{}.
\newblock \bibinfo{title}{{LLM}-grounded {Diffusion}: {Enhancing} {Prompt}
  {Understanding} of {Text}-to-{Image} {Diffusion} {Models} with {Large}
  {Language} {Models}}.
\newblock
\newblock
\urldef\tempurl%
\url{http://arxiv.org/abs/2305.13655}
\showURL{%
\tempurl}
\newblock
\shownote{arXiv:2305.13655 [cs]}.


\bibitem[Liu et~al\mbox{.}(2023)]%
        {liu_visual_2023}
\bibfield{author}{\bibinfo{person}{Haotian Liu}, \bibinfo{person}{Chunyuan Li},
  \bibinfo{person}{Qingyang Wu}, {and} \bibinfo{person}{Yong~Jae Lee}.}
  \bibinfo{year}{2023}\natexlab{}.
\newblock \bibinfo{title}{Visual {Instruction} {Tuning}}.
\newblock
\newblock
\urldef\tempurl%
\url{https://doi.org/10.48550/arXiv.2304.08485}
\showDOI{\tempurl}
\newblock
\shownote{arXiv:2304.08485 [cs]}.


\bibitem[Liu et~al\mbox{.}(2022)]%
        {liu_compositional_2022}
\bibfield{author}{\bibinfo{person}{Nan Liu}, \bibinfo{person}{Shuang Li},
  \bibinfo{person}{Yilun Du}, \bibinfo{person}{Antonio Torralba}, {and}
  \bibinfo{person}{Joshua~B. Tenenbaum}.} \bibinfo{year}{2022}\natexlab{}.
\newblock \bibinfo{title}{Compositional {Visual} {Generation} with {Composable}
  {Diffusion} {Models}}.
\newblock
\newblock
\urldef\tempurl%
\url{https://arxiv.org/abs/2206.01714v6}
\showURL{%
\tempurl}


\bibitem[Min et~al\mbox{.}(2022)]%
        {min_metaicl_2022}
\bibfield{author}{\bibinfo{person}{Sewon Min}, \bibinfo{person}{Mike Lewis},
  \bibinfo{person}{Luke Zettlemoyer}, {and} \bibinfo{person}{Hannaneh
  Hajishirzi}.} \bibinfo{year}{2022}\natexlab{}.
\newblock \bibinfo{title}{{MetaICL}: {Learning} to {Learn} {In} {Context}}.
\newblock
\newblock
\urldef\tempurl%
\url{https://doi.org/10.48550/arXiv.2110.15943}
\showDOI{\tempurl}
\newblock
\shownote{arXiv:2110.15943 [cs]}.


\bibitem[Parmar et~al\mbox{.}(2023)]%
        {parmar_zero-shot_2023}
\bibfield{author}{\bibinfo{person}{Gaurav Parmar},
  \bibinfo{person}{Krishna~Kumar Singh}, \bibinfo{person}{Richard Zhang},
  \bibinfo{person}{Yijun Li}, \bibinfo{person}{Jingwan Lu}, {and}
  \bibinfo{person}{Jun-Yan Zhu}.} \bibinfo{year}{2023}\natexlab{}.
\newblock \bibinfo{title}{Zero-shot {Image}-to-{Image} {Translation}}.
\newblock
\newblock
\urldef\tempurl%
\url{https://doi.org/10.48550/arXiv.2302.03027}
\showDOI{\tempurl}
\newblock
\shownote{arXiv:2302.03027 [cs]}.


\bibitem[Perrault and Grosz(1988)]%
        {perrault_chapter_1988}
\bibfield{author}{\bibinfo{person}{C.~Raymond Perrault} {and}
  \bibinfo{person}{Barbara~J. Grosz}.} \bibinfo{year}{1988}\natexlab{}.
\newblock \showarticletitle{Chapter 4 - {Natural}-{Language} {Interfaces}}.
\newblock In \bibinfo{booktitle}{\emph{Exploring {Artificial} {Intelligence}}},
  \bibfield{editor}{\bibinfo{person}{Howard~E. Shrobe} {and}
  \bibinfo{person}{{the American Association for Artificial Intelligence}}}
  (Eds.). \bibinfo{publisher}{Morgan Kaufmann}, \bibinfo{pages}{133--172}.
\newblock
\showISBNx{978-0-934613-67-5}
\urldef\tempurl%
\url{https://doi.org/10.1016/B978-0-934613-67-5.50008-3}
\showDOI{\tempurl}


\bibitem[Ramesh et~al\mbox{.}(2021)]%
        {ramesh_zero-shot_2021}
\bibfield{author}{\bibinfo{person}{Aditya Ramesh}, \bibinfo{person}{Mikhail
  Pavlov}, \bibinfo{person}{Gabriel Goh}, \bibinfo{person}{Scott Gray},
  \bibinfo{person}{Chelsea Voss}, \bibinfo{person}{Alec Radford},
  \bibinfo{person}{Mark Chen}, {and} \bibinfo{person}{Ilya Sutskever}.}
  \bibinfo{year}{2021}\natexlab{}.
\newblock \bibinfo{title}{Zero-{Shot} {Text}-to-{Image} {Generation}}.
\newblock
\newblock
\urldef\tempurl%
\url{https://doi.org/10.48550/arXiv.2102.12092}
\showDOI{\tempurl}
\newblock
\shownote{arXiv:2102.12092 [cs]}.


\bibitem[Rombach et~al\mbox{.}(2022)]%
        {rombach_high-resolution_2022}
\bibfield{author}{\bibinfo{person}{Robin Rombach}, \bibinfo{person}{Andreas
  Blattmann}, \bibinfo{person}{Dominik Lorenz}, \bibinfo{person}{Patrick
  Esser}, {and} \bibinfo{person}{Björn Ommer}.}
  \bibinfo{year}{2022}\natexlab{}.
\newblock \bibinfo{title}{High-{Resolution} {Image} {Synthesis} with {Latent}
  {Diffusion} {Models}}.
\newblock
\newblock
\urldef\tempurl%
\url{https://doi.org/10.48550/arXiv.2112.10752}
\showDOI{\tempurl}
\newblock
\shownote{arXiv:2112.10752 [cs]}.


\bibitem[Ruiz et~al\mbox{.}(2023)]%
        {ruiz_dreambooth_2023}
\bibfield{author}{\bibinfo{person}{Nataniel Ruiz}, \bibinfo{person}{Yuanzhen
  Li}, \bibinfo{person}{Varun Jampani}, \bibinfo{person}{Yael Pritch},
  \bibinfo{person}{Michael Rubinstein}, {and} \bibinfo{person}{Kfir Aberman}.}
  \bibinfo{year}{2023}\natexlab{}.
\newblock \bibinfo{title}{{DreamBooth}: {Fine} {Tuning} {Text}-to-{Image}
  {Diffusion} {Models} for {Subject}-{Driven} {Generation}}.
\newblock
\newblock
\urldef\tempurl%
\url{https://doi.org/10.48550/arXiv.2208.12242}
\showDOI{\tempurl}
\newblock
\shownote{arXiv:2208.12242 [cs]}.


\bibitem[Saharia et~al\mbox{.}(2022)]%
        {saharia_photorealistic_2022}
\bibfield{author}{\bibinfo{person}{Chitwan Saharia}, \bibinfo{person}{William
  Chan}, \bibinfo{person}{Saurabh Saxena}, \bibinfo{person}{Lala Li},
  \bibinfo{person}{Jay Whang}, \bibinfo{person}{Emily Denton},
  \bibinfo{person}{Seyed Kamyar~Seyed Ghasemipour},
  \bibinfo{person}{Burcu~Karagol Ayan}, \bibinfo{person}{S.~Sara Mahdavi},
  \bibinfo{person}{Rapha~Gontijo Lopes}, \bibinfo{person}{Tim Salimans},
  \bibinfo{person}{Jonathan Ho}, \bibinfo{person}{David~J. Fleet}, {and}
  \bibinfo{person}{Mohammad Norouzi}.} \bibinfo{year}{2022}\natexlab{}.
\newblock \bibinfo{title}{Photorealistic {Text}-to-{Image} {Diffusion} {Models}
  with {Deep} {Language} {Understanding}}.
\newblock
\newblock
\urldef\tempurl%
\url{https://doi.org/10.48550/arXiv.2205.11487}
\showDOI{\tempurl}
\newblock
\shownote{arXiv:2205.11487 [cs]}.


\bibitem[Schramowski et~al\mbox{.}(2023)]%
        {schramowski_safe_2023}
\bibfield{author}{\bibinfo{person}{Patrick Schramowski},
  \bibinfo{person}{Manuel Brack}, \bibinfo{person}{Björn Deiseroth}, {and}
  \bibinfo{person}{Kristian Kersting}.} \bibinfo{year}{2023}\natexlab{}.
\newblock \showarticletitle{Safe {Latent} {Diffusion}: {Mitigating}
  {Inappropriate} {Degeneration} in {Diffusion} {Models}}.
  \bibinfo{pages}{22522--22531}.
\newblock
\urldef\tempurl%
\url{https://openaccess.thecvf.com/content/CVPR2023/html/Schramowski_Safe_Latent_Diffusion_Mitigating_Inappropriate_Degeneration_in_Diffusion_Models_CVPR_2023_paper.html}
\showURL{%
\tempurl}


\bibitem[Schuhmann et~al\mbox{.}(2022)]%
        {schuhmann2022laion5b}
\bibfield{author}{\bibinfo{person}{Christoph Schuhmann},
  \bibinfo{person}{Romain Beaumont}, \bibinfo{person}{Richard Vencu},
  \bibinfo{person}{Cade Gordon}, \bibinfo{person}{Ross Wightman},
  \bibinfo{person}{Mehdi Cherti}, \bibinfo{person}{Theo Coombes},
  \bibinfo{person}{Aarush Katta}, \bibinfo{person}{Clayton Mullis},
  \bibinfo{person}{Mitchell Wortsman}, \bibinfo{person}{Patrick Schramowski},
  \bibinfo{person}{Srivatsa Kundurthy}, \bibinfo{person}{Katherine Crowson},
  \bibinfo{person}{Ludwig Schmidt}, \bibinfo{person}{Robert Kaczmarczyk}, {and}
  \bibinfo{person}{Jenia Jitsev}.} \bibinfo{year}{2022}\natexlab{}.
\newblock \bibinfo{title}{LAION-5B: An open large-scale dataset for training
  next generation image-text models}.
\newblock
\newblock
\showeprint[arxiv]{2210.08402}~[cs.CV]


\bibitem[Sheynin et~al\mbox{.}(2023)]%
        {sheynin_emu_2023}
\bibfield{author}{\bibinfo{person}{Shelly Sheynin}, \bibinfo{person}{Adam
  Polyak}, \bibinfo{person}{Uriel Singer}, \bibinfo{person}{Yuval Kirstain},
  \bibinfo{person}{Amit Zohar}, \bibinfo{person}{Oron Ashual},
  \bibinfo{person}{Devi Parikh}, {and} \bibinfo{person}{Yaniv Taigman}.}
  \bibinfo{year}{2023}\natexlab{}.
\newblock \bibinfo{title}{Emu {Edit}: {Precise} {Image} {Editing} via
  {Recognition} and {Generation} {Tasks}}.
\newblock
\newblock
\urldef\tempurl%
\url{https://doi.org/10.48550/arXiv.2311.10089}
\showDOI{\tempurl}
\newblock
\shownote{arXiv:2311.10089 [cs]}.


\bibitem[Shneiderman(1980)]%
        {shneiderman_1980}
\bibfield{author}{\bibinfo{person}{Ben Shneiderman}.}
  \bibinfo{year}{1980}\natexlab{}.
\newblock \showarticletitle{Natural vs. precise concise languages for human
  operation of computers: research issues and experimental approaches}. In
  \bibinfo{booktitle}{\emph{Proceedings of the 18th Annual Meeting on
  Association for Computational Linguistics}} (Philadelphia, Pennsylvania)
  \emph{(\bibinfo{series}{ACL '80})}. \bibinfo{publisher}{Association for
  Computational Linguistics}, \bibinfo{address}{USA},
  \bibinfo{pages}{139–141}.
\newblock
\urldef\tempurl%
\url{https://doi.org/10.3115/981436.981478}
\showDOI{\tempurl}


\bibitem[SHNEIDERMAN(1982)]%
        {shneiderman_future_1982}
\bibfield{author}{\bibinfo{person}{BEN SHNEIDERMAN}.}
  \bibinfo{year}{1982}\natexlab{}.
\newblock \showarticletitle{The future of interactive systems and the emergence
  of direct manipulation†}.
\newblock \bibinfo{journal}{\emph{Behaviour \& Information Technology}}
  \bibinfo{volume}{1}, \bibinfo{number}{3} (\bibinfo{date}{July}
  \bibinfo{year}{1982}), \bibinfo{pages}{237--256}.
\newblock
\showISSN{0144-929X}
\urldef\tempurl%
\url{https://doi.org/10.1080/01449298208914450}
\showDOI{\tempurl}
\newblock
\shownote{Publisher: Taylor \& Francis \_eprint:
  https://doi.org/10.1080/01449298208914450}.


\bibitem[Shneiderman(1983)]%
        {shneiderman1983direct}
\bibfield{author}{\bibinfo{person}{Ben Shneiderman}.}
  \bibinfo{year}{1983}\natexlab{}.
\newblock \showarticletitle{Direct manipulation: A step beyond programming
  languages}.
\newblock \bibinfo{journal}{\emph{Computer}} \bibinfo{volume}{16},
  \bibinfo{number}{08} (\bibinfo{year}{1983}), \bibinfo{pages}{57--69}.
\newblock


\bibitem[Shneiderman and Maes(1997)]%
        {shneiderman1997direct}
\bibfield{author}{\bibinfo{person}{Ben Shneiderman} {and}
  \bibinfo{person}{Pattie Maes}.} \bibinfo{year}{1997}\natexlab{}.
\newblock \showarticletitle{Direct manipulation vs. interface agents}.
\newblock \bibinfo{journal}{\emph{interactions}} \bibinfo{volume}{4},
  \bibinfo{number}{6} (\bibinfo{year}{1997}), \bibinfo{pages}{42--61}.
\newblock


\bibitem[van Dam(1997)]%
        {van_dam_post-wimp_1997}
\bibfield{author}{\bibinfo{person}{Andries van Dam}.}
  \bibinfo{year}{1997}\natexlab{}.
\newblock \showarticletitle{Post-{WIMP} user interfaces}.
\newblock \bibinfo{journal}{\emph{Commun. ACM}} \bibinfo{volume}{40},
  \bibinfo{number}{2} (\bibinfo{date}{Feb.} \bibinfo{year}{1997}),
  \bibinfo{pages}{63--67}.
\newblock
\showISSN{0001-0782}
\urldef\tempurl%
\url{https://doi.org/10.1145/253671.253708}
\showDOI{\tempurl}


\bibitem[Wang et~al\mbox{.}(2023)]%
        {wang2023diffusiondb}
\bibfield{author}{\bibinfo{person}{Zijie~J. Wang}, \bibinfo{person}{Evan
  Montoya}, \bibinfo{person}{David Munechika}, \bibinfo{person}{Haoyang Yang},
  \bibinfo{person}{Benjamin Hoover}, {and} \bibinfo{person}{Duen~Horng Chau}.}
  \bibinfo{year}{2023}\natexlab{}.
\newblock \bibinfo{title}{DiffusionDB: A Large-scale Prompt Gallery Dataset for
  Text-to-Image Generative Models}.
\newblock
\newblock
\showeprint[arxiv]{2210.14896}~[cs.CV]


\bibitem[Wei et~al\mbox{.}(2023)]%
        {wei_chain--thought_2023}
\bibfield{author}{\bibinfo{person}{Jason Wei}, \bibinfo{person}{Xuezhi Wang},
  \bibinfo{person}{Dale Schuurmans}, \bibinfo{person}{Maarten Bosma},
  \bibinfo{person}{Brian Ichter}, \bibinfo{person}{Fei Xia},
  \bibinfo{person}{Ed Chi}, \bibinfo{person}{Quoc Le}, {and}
  \bibinfo{person}{Denny Zhou}.} \bibinfo{year}{2023}\natexlab{}.
\newblock \bibinfo{title}{Chain-of-{Thought} {Prompting} {Elicits} {Reasoning}
  in {Large} {Language} {Models}}.
\newblock
\newblock
\urldef\tempurl%
\url{https://doi.org/10.48550/arXiv.2201.11903}
\showDOI{\tempurl}
\newblock
\shownote{arXiv:2201.11903 [cs]}.


\end{thebibliography}

\end{document}